\documentclass{article}

\usepackage{PRIMEarxiv}

\usepackage[utf8]{inputenc} 
\usepackage[T1]{fontenc}    
\usepackage{hyperref}       
\usepackage{url}            
\usepackage{booktabs}       
\usepackage{amsfonts}       
\usepackage{nicefrac}       
\usepackage{microtype}      
\usepackage{lipsum}
\usepackage{fancyhdr}       
\usepackage{graphicx}       
\usepackage{caption}
\usepackage{subcaption}
\usepackage[ruled,vlined]{algorithm2e}
\usepackage{color}
\usepackage{multirow}
\usepackage[font=small,labelfont=bf]{caption}
\graphicspath{{media/}}     

\title{BERT WEAVER: Using WEight AVERaging to enable lifelong learning for transformer-based models in biomedical semantic search engines
}

\author{
  Lisa Kühnel \\
  \textit{ZB MED - Information Centre for Life Sciences} \\
  Cologne, Germany\\
  \textit{Graduate School DILS, Bielefeld Institute for Bioinformatics Infrastructure (BIBI),}\\
  \textit{Faculty of Technology, Bielefeld University}  \\
  Bielefeld, Germany \\
  kuehnel@zbmed.de \\
   \And
  Alexander Schulz \\
    \textit{CITEC, Bielefeld University} \\
    Bielefeld, Germany \\
    aschulz@techfak.uni-bielefeld.de \\
   \And
   Barbara Hammer \\
\textit{CITEC, Bielefeld University} \\
    Bielefeld, Germany \\
    bhammer@techfak.uni-bielefeld.de \\
   \And
   Juliane Fluck \\
\textit{ZB MED - Information Centre for Life Sciences} \\
   Cologne, Germany \\
   \textit{University of Bonn} \\
   Bonn, Germany \\
   fluck@zbmed.de \\
}

\begin{document}
\maketitle

\begin{abstract}
   Recent developments in transfer learning have boosted the advancements in natural language processing tasks. The performance is, however, dependent on high-quality, manually annotated training data. Especially in the biomedical domain, it has been shown that one training corpus is not enough to learn generic models that are able to efficiently predict on new data. Therefore, in order to be used in real world applications state-of-the-art models need the ability of lifelong learning to improve performance as soon as new data are available - without the need of re-training the whole model from scratch. We present WEAVER, a simple, yet efficient post-processing method that infuses old knowledge into the new model, thereby reducing catastrophic forgetting. We show that applying WEAVER in a sequential manner results in similar word embedding distributions as doing a combined training on all data at once, while being computationally more efficient. Because there is no need of data sharing, the presented method is also easily applicable to federated learning settings and can for example be beneficial for the mining of electronic health records from different clinics.
\end{abstract}


The amount of literature in the medical domain is increasing enormously and emphasizes the need for text mining-based solutions in order to automatically extract relevant information. Named entity recognition (NER) is an important task of natural language processing (NLP) where the aim is to find entity classes in unstructured text, such as specific diseases. As the amount of data for a specific setup is usually limited, recently, transfer learning-based models have been shown to achieve state-of-the-art results in many NLP tasks including NER \cite{yadav2019survey}. Especially, transformer-based models, such as BERT \cite{DBLP:conf/naacl/DevlinCLT19}, show promising results on benchmark tasks \cite{Kulkarni2022}. In the biomedical domain, BioBERT \cite{lee_biobert_2020} shows state-of-the-art performance for several NER tasks, such as disease recognition. Promising F1-scores are achieved for the available data sets (above 84\%). 

\par Based on the use case of disease NER, we recently showed that models trained on an available data set are not able to efficiently predict on another data set that, however, follows the same annotation guidelines \cite{kuhnel_we_2022}. This is not only true for transformer-based models such as BioBERT but holds true for different machine learning-based models such as convolutional neural networks or conditional random fields. In our previous study, we showed -- based on five different manually labeled data sets -- that the performance of a model trained on one of these corpora is reduced by up to 20\% in terms of F1-score when predicting on another corpus. This significant drop in performance indicates that the training data is either too small or not representative - compared to a random PubMed corpus. One reason can be attributed to the fact that specific corpora are often comparably small such that small differences in between those data sets are mapped to differences in embeddings and according NER downstream tasks. 
Therefore, in order to use these models in real world applications, such as semantic search engines, it is advisable to improve the models as soon as new annotated data are available, to obtain optimum performance. This process is known as lifelong learning or, equivalently, continual learning (CL), which means that a model is sequentially improved in a so-called online fashion \cite{DBLP:journals/corr/abs-1912-05156}. However, for such settings, a mechanism called catastrophic forgetting easily happens \cite{mccloskey_catastrophic_1989}. This means that the model will be biased towards the last data set and will forget previously learned structures. Therefore, the aim of continual learning algorithms is to minimize catastrophic forgetting. Besides, knowledge transfer (KT) should happen, which means that both already learned knowledge helps to learn a new task and newly acquired knowledge improves the performance on the old task, known as forward- and backward transfer, respectively.

In the beginning of the COVID-19 pandemic, we developed a semantic search engine in order to help information specialists and researchers to cope with the high amount of publications arising daily -- especially in the form of preprints. In this service, several different text mining-based components are integrated, such as for the recognition of disease names, virus proteins or the classification of long COVID related documents. Moreover, we integrated a feedback button, where users (who are mainly information specialists) can give us direct feedback on the annotations~\cite{langnickel_continuous_2022}. In line with our previous research \cite{kuhnel_we_2022}, we realized several performance drops in the real world applications as compared to the available test sets used for evaluation of the algorithms. Therefore, user feedback is of great importance so that we can generate new data sets and improve our included models. To do this efficiently, a lifelong learning capacity is essential.

In the literature, several subcategories of continual learning are discussed that require different network architectures. According to \cite{ke2023continual}, there is task-incremental learning (TIL) that aims to learn a series of different tasks, such as named entity recognition and document classification. In a typical TIL-setting, a task-id is given and new weights (output layers) are parameterized for each new task. Furthermore, class-incremental learning (CIL) exists, where non-overlapping classes (e.g. diseases, genes, etc.) are learned continually for the same task; per definition, no task-id is provided at test time. In domain-incremental learning (DIL) the same task is learned, but for different domains. A last -- and less well studied -- category is known as temporal CL, where only one task from one domain is learned that needs to be updated as the time goes, e.g. due to temporal changes in language.

As described before, the integration of text mining components into running services in the biomedical domain suffers from (1) very specific training corpora that can be specific towards specific medical areas, e.g. drug interactions and (2) temporal changes in language, such as the emergence of new diseases. Hence, our problem does not directly belong to one of the mentioned categories but can be interpreted as a combination of domain-incremental learning (on a fine-granular level) and temporal CL.

Therefore, we developed a new continual learning method -- called WEAVER -- that can be applied to transformer-based models which exploits parts of the federated averaging algorithm (FedAvg), known from federated learning approaches~\cite{mcmahan_communication-efficient_2017}. Thereby, previously used data are not required for the new task, the model structure does not need to be changed and the process is computationally efficient. For the real world applications described above, the model can hence be efficiently re-trained as soon as new data are available.

We formally define our research problem in Chapter~\ref{sec:problem_def} and describe related work in Chapter \ref{sec:sota_cl}. Afterwards, our proposed algorithm is described in Chapter \ref{sec:weaver}. Performed experiments, the results and the discussion can be found in Chapters \ref{sec:experiments}, \ref{sec:results} and \ref{sec:discussion}, respectively. We conclude our findings in Chapter \ref{sec:conclusion}.

\section{Problem Definition}\label{sec:problem_def}
The learning task can be described as follows. Given a sequence of different labeled data sets $D = (D_1, D_2, D_3, ... , D_n)$ for the same task $T$ and the same class $C$, we want to learn from them sequentially without forgetting what has been learned previously. Thereby, the data sets may or may not be disjoint -- depending on what the users annotated -- and may be of different sizes. Note, that in our setting, a training task refers to both the same task (i.e. NER) and entity class, e.g. diseases, and therefore does not directly belong to one of the four mentioned CL-settings.

\par This results in the following requirements: We need a method that is able to cope with different data set sizes. Moreover, it needs to be computationally efficient, especially at inference time, to ensure reasonable processing time for the integration into the service. In line with this, we need a non-varying model architecture to ensure constant memory requirements and constant speed at inference time. Finally, we have the constraint that the method does not rely on seeing previous data sets $(D_1, D_2, D_3, ... , D_{n-1})$ (hence, data sets are not stored) because (a) a re-training from scratch is computationally inefficient, and (b) this ensures that the method can also be applied to medical use cases where sharing of data sets is not allowed due to privacy regulations.

\section{Related Work}\label{sec:sota_cl}
Several overview articles structure and compare online learning methods and their suitability in various domains, e.g. \cite{HOI2021249,DBLP:journals/ijon/LosingHW18}. This section summarizes available state-of-the-art methods for different continual learning scenarios with a focus on the application to NLP use cases.
\par A lot of research has been done in the area of continual learning to prevent a model from forgetting. One of the most prominent approaches is called Elastic Weight Consolidation (EWC) proposed by Kirkpatrick \textit{et al.} \cite{kirkpatrick_overcoming_2017}. It is a regularization-based technique that basically quantifies the importance of weights and thereby impedes important weights from being changed drastically. It has been successfully applied for an online personalization of speech recognition systems, as an example \cite{sim2019personalization}. Based on EWC, Liu \textit{et al.} proposed an extension that makes use of a network reparameterization that basically rotates the parameter space \cite{liu_rotate_2018}. More recently, Aljundi \textit{et al.} proposed Memory Aware Synapses (MAS) -- a method that, given a new sample, measures the importance of each parameter of the network by taking the sensitivity of the predicted output towards a change in that parameter into account \cite{aljundi_memory_2018}. Several examples also exist in the NLP domain. For example, Li \textit{et al.} proposed an algorithm, called RMR-DSE, standing for regularization memory recall -- domain shift estimation, contains a modified version of the EWC algorithm~\cite{li_overcoming_2022}. Thereby, instead of using the Fisher Information Matrix as criterion to judge on the importance of each weight based on the previously seen data, they introduce a learnable parameter to solve this task that is standardized by a further hyperparameter that takes the size of the vocabularies for each respective data set into account. The second component, the domain shift estimation part, compares embeddings from the current and previous model and clusters them to calculate semantic shifts and adjust the weights accordingly. This mechanism is also used at inference time by comparing the input test sentences to previously stored embeddings from the training data. Therefore, this methods is not feasible for the integration into a running service because of its inefficiency at inference.

\par Next to regularization-based techniques, (pseudo-)rehearsal-based approaches have been proposed, e.g. \cite{robins_catastrophic_1995, honnibal_pseudo-rehearsal_nodate}. Rehearsal means that a subset of previously seen data is combined with the new data. Since the old data are not always available, these methods often include a generator network that generates new data based on the previously seen data set (e.g. \cite{qin2022lfpt5}). This is also often called silver standard or replay buffer. These data are then mixed with new data to re-train the model. For rehearsal-based methods, research has been done on how best to select the replay buffer for an efficient training, e.g. gradient-based selection has been proposed, known under the abbreviation GEM -- Gradient Episodic Memory -- where several algorithms and extensions have been proposed recently, e.g. \cite{aljundi_gradient_2019, lopez-paz_gradient_2022, chaudhry_efficient_2023}. Experience replay is another rehearsal-based approach, for example investigated by~\cite{rostami_complementary_2019}. Further transformer-based methods include Memory-based parameter adaptation ($MbPA^{++}$) \cite{dautume_episodic_2019}, Language Modeling for Lifelong Language Learning (LAMOL) \cite{sun_lamol_2019}, and its extension Lifelong Learning Knowledge Distillation (L2KD) \cite{chuang_lifelong_2020}. Whereas the two latter simultaneously train a \emph{learner} and a \emph{generator} network, $MbPA^{++}$ uses sparse experience replay and local adaptation. Due to our constraint that the method should be applicable to medical use cases as well where a sharing of data sets is not possible, rehearsal-based methods cannot be applied to our use case.

\par Finally, promising methods exist where new parameters are added to the model for each new task that is learned, such as proposed by Fayek \textit{et al.} \cite{fayek_progressive_2020}. A well known example from the transformer domain is called AdapterFusion~\cite{pfeiffer_adapterfusion_2021}. Adapters, proposed by Houlsby \textit{et al.} are transformer-based modules exploiting a more parameter-efficient learning procedure than the usual fine-tuning \cite{houlsby_parameter-efficient_2019}. Except from being more parameter-efficient, these Adapters can be used for continual learning settings as they can be trained individually and then be fused together based on the target (i.e. the final) task in a knowledge composition step. Zhang \textit{et al.} build upon Adapters, but instead of training a new module for each task, the authors developed an algorithm that decides whether a completely new adapter needs to be trained or whether an already trained one can be re-used~\cite{zhang_continual_2022}. In addition, they apply pseudo experience replay according to~\cite{sun_lamol_2019}. Moreover, dual-memory-based methods are applied where two different networks are used - one for memorising already learned information and one for learning new tasks - such as shown by Hattori~\cite{hattori_biologically_2014} or Park~\cite{park_continual_2020}. The architecture proposed by Hattori is strongly inspired by biological processes in the brain, making use of so-called hippocampal and neonortical networks. In contrast, Park implemented a dual network architecture based on state-of-the-art transformers. As defined in Section~\ref{sec:problem_def}, methods using task-specific model components or different network architectures for different tasks/data sets cannot be used for the integration into our service due to undefined memory requirements and the inability of ensuring a consistent inference time.

Because of our constraint to not use old data while re-training on a new data set, it might be meaningful to have a look at current state-of-the-art-methods in federated learning settings, because these settings have indeed the definition that the data sets are stored at different places. Recently proposed approaches in the NLP-domain, such as~\cite{liu2020federated}, include local fine-tuning of individual pre-trained transformer models, which are then sent to a central server where weight averaging is performed according to McMahan \textit{et al.}~\cite{mcmahan_communication-efficient_2017}. This is performed iteratively, such that after each epoch, the weights are sent to the server and updated before sending them back to the individual locations. This approach, however, requires the data sets to be at the same place, which is not given in a continual learning setting.

\par Consequently, in this work, we propose a new continual learning method that fulfills our requirements for the integration into a running service.

\section{Proposed Algorithm}\label{sec:weaver}

\begin{figure*}
    \centering
    \includegraphics[width=0.7\linewidth]{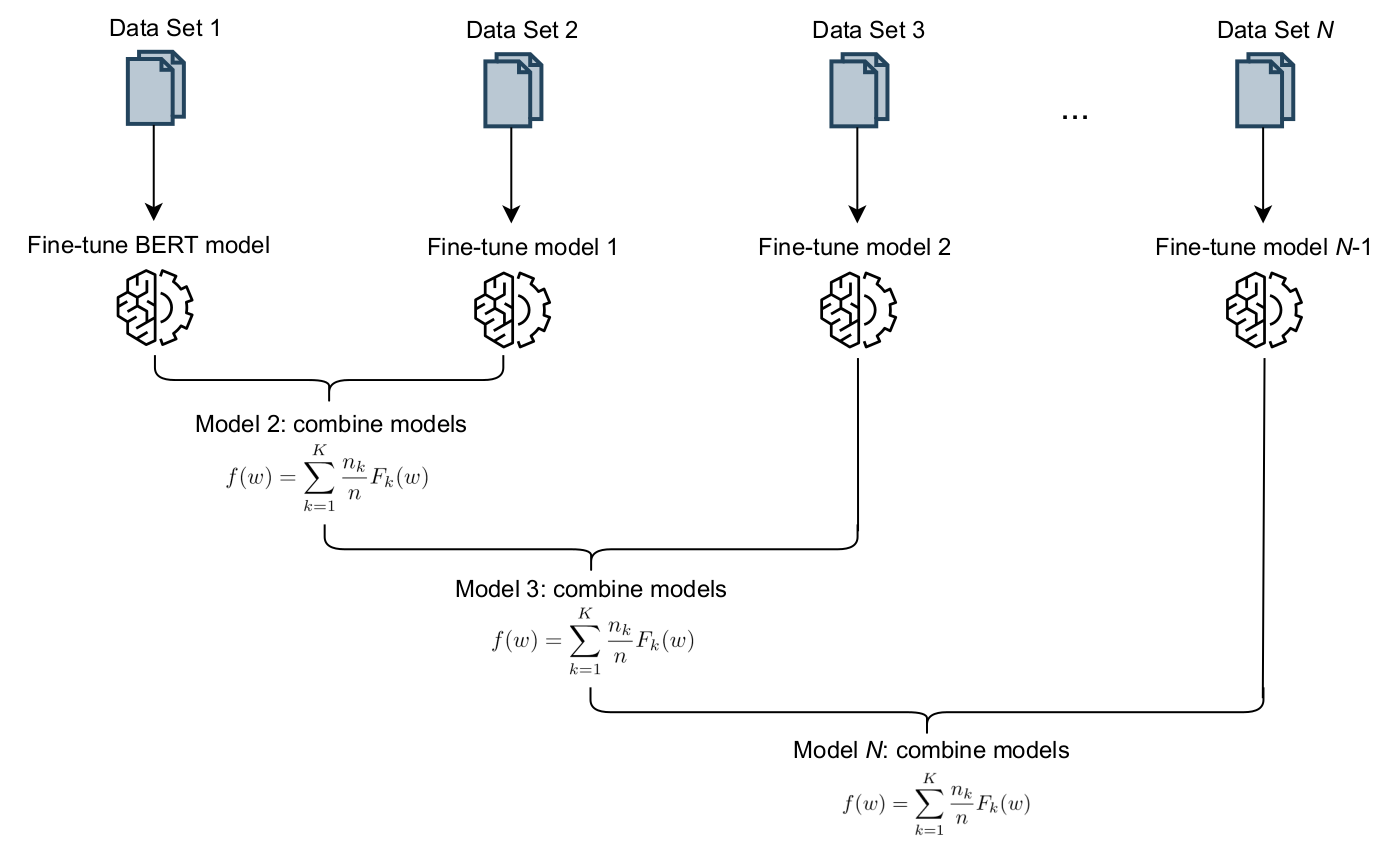}
    \caption{\textbf{Overview of our transformer-based continual learning procedure WEAVER using weight averaging.} For training the first model, a transformer-based model, such as BERT, is initialized and fine-tuned on the available data. To continue training in a sequential manner, the already fine-tuned model is fine-tuned again on a second corpus. To prevent catastrophic forgetting, knowledge of the previous model is infused into the new model by applying weight averaging. Thereby, the size of the data set the individual model was trained on determines the averaging coefficient, i.e. the bigger the data set, the higher the influence for the new model. This procedure is repeated for every new data set.}
    \label{fig:overview}
\end{figure*}

For our proposed continual learning procedure, we exploit a mechanism that is originally used in federated learning settings, where models are trained at different places using different data sets, mostly due to data privacy concerns \cite{DBLP:journals/corr/abs-1907-09693}. After training these models individually, their weights are passed to a central server and averaged in relation to the amount of training data they were trained on - hence, the more data were available the more influence in the final model \cite{mcmahan_communication-efficient_2017}. The corresponding formula for the objective of the target model can be seen in the following:

\begin{equation} \label{eq:fedAvg}
f(w) = \sum_{k=1}^{K} \frac{n_k}{n} F_k(w)
\end{equation}
$K$ is the number of clients, i.e. the number of models that were trained. The total amount of training data is described by $n$, whereas $n_k$ is the amount of the current data set. $F_k(w)$ defines the client's loss function. As shown in \cite{mcmahan_communication-efficient_2017}, this objective results in weight averaging for convex costs.
\newline
Based on this, we developed the following procedure: For the first model that is trained on a given task, we initialize a pre-trained BioBERT model and fine-tune it in a usual manner. As soon as new data are then available, we fine-tune the already trained model again using the new data set. In a post-processing step, the weights of the old and the new model are then averaged, taking the amount of training data into account. Thereby, if a second model is trained on top of the first one, the total amount of training data is the sum of the two data sets. Therefore, either a new pre-trained model can be initialized or the already fine-tuned model will be fine-tuned again and afterwards combined. A simplified overview about the continual learning procedure is shown in Fig.~\ref{fig:overview} and we provide the pseudo-code in the following.

\SetKwInput{KwInput}{Input}                
\SetKwInput{KwOutput}{Output}              

\begin{algorithm}
\SetAlgoLined
  \KwInput{Pre-trained BERT model, Data sets $D = (D_1, ... , D_n)$}
  \KwOutput{Continually fine-tuned BERT model}
 \For{$D_i$ in $D$}{
  curr\_model= Train(model, $D_i$)\;
  \If {indexOf($D_i$) != 0}
      {
      curr\_data = size($D_i$)\;
      all\_data += curr\_data\;
      updated\_model = WeightAverage(model, curr\_model, all\_data, curr\_data)\;
      model = updated\_model\;
      }
  \Else{model = curr\_model\;
  all\_data = size($D_i$)\;}
 }
 \caption{Continual Learning with WEAVER}\label{pseudo:weaver}
\end{algorithm}

\section{Experiments}\label{sec:experiments}
In this section, we first describe the used data sets. Afterwards, the conducted experiments and their implementation details are given. Finally, we describe our evaluation and visualization strategies in detail. 

\subsection{Used Data Sets} \label{sec:datasets}
We perform the experiments on three different NER tasks from the biomedical domain. For disease NER, we use five different data sets - four of them have been described in detail in our previous publication \cite{kuhnel_we_2022}. Additionally, we use the plant-disease corpus \cite{kim_corpus_2019}. For both proteins/genes and chemicals, we rely on six and five different datasets, respectively, described in detail by Crichton \textit{et al.} \cite{crichton_neural_2017} and provided under \url{https://github.com/cambridgeltl/MTL-Bioinformatics-2016}. An overview of the used data sets can be seen in Table~\ref{tab:overview_datasets}.

\begin{table}
    \centering
    \caption{Overview of used datasets from the biomedical domain}
    \begin{tabular}{lll}
         NER-Task & Dataset & Size* \\ 
         \hline
         \multirow{5}{*}{Disease} & NCBI \cite{dogan_ncbi_2014} & 4725 \\
         & BC5CDR \cite{lee_biobert_2020} & 3230 \\
         & miRNA-disease \cite{bagewadi_detecting_2015} & 3043 \\ 
         & plant-disease \cite{kim_corpus_2019} & 2944 \\ 
         & BioNLP13-CG \cite{pyysalo_overview_2013} & 1885 \\ 
         \hline
         \multirow{6}{*}{Proteins/Genes} & BioNLP11-ID \cite{pyysalo_overview_2012} & 3197 \\
         &  BioNLP13-CG \cite{pyysalo_overview_2013} &  4027 \\
         &  BioNLP13-GE \cite{kim_genia_nodate} & 3566 \\
         &  BioNLP13-PC \cite{ohta_overview_2013} & 5468 \\
         &  Ex-PTM \cite{pyysalo_towards_2011} & 1787 \\ 
         &  JNLPBA \cite{kim_introduction_2004} & 46750 \\
         \hline
         \multirow{5}{*}{Chemicals} & BioNLP13-CG \cite{pyysalo_overview_2013} & 1097 \\
         &  BioNLP13-PC \cite{ohta_overview_2013} & 1178 \\
         &  BC4CHEMD \cite{krallinger_chemdner_2015} & 29478 \\
         &  BioNLP11-ID \cite{pyysalo_overview_2012} & 594 \\ 
         &  BC5CDR \cite{lee_biobert_2020} & 5203 \\
         \hline
         \multicolumn{3}{l}{*Size refers to the amount of entities in the training set.}
    \end{tabular}
    \label{tab:overview_datasets}
\end{table}

We simulate a continuous learning setting, where different datasets are learned sequentially. As in real-world settings, the datasets can differ in size and slightly differ in used annotation guidelines. According to \cite{dautume_episodic_2019}, we randomly chose four different orders for each setup (as can be seen in Table~\ref{tab:orders}).

\subsection{Conducted Experiments}
As baseline experiments, we train one model individually on each of the available data sets. Each model is then evaluated on all available test data sets for this entity class. For example, a model trained on the NCBI training corpus is evaluated on the corresponding test set but also on the four other test sets (BC5CDR, BioNLP13, miRNA-disease and plant-disease).

We perform the following CL-based experiments to evaluate and compare our developed algorithm.

\begin{itemize}
    \item \textbf{FineTune}: a standard BERT model that is fine tuned sequentially for each new data set
    \item \textbf{EWC} \cite{kirkpatrick_overcoming_2017}: our own implementation of EWC for NER with transformer-based models
    \item \textbf{AdapterFusion} \cite{houlsby_parameter-efficient_2019}: one adapter is individually trained per training data set and they are sequentially fused together
    \item \textbf{WEAVER}: our model described in Chapter \ref{sec:weaver}
    \item \textbf{Replay}: Sparse experience replay mechanism as performed by \cite{dautume_episodic_2019}. While fine-tuning BERT models sequentially, we replay 10\% of all previously seen data.
    \item \textbf{MTL}: a multi-task upper-bound model trained on all available training data sets simultaneously
\end{itemize}

Note, that Replay and MTL require the datasets to be at one place and therefore just serve as upper-bound methods for comparative reasons in our study. For the same reason, we omit other state-of-the-art transformer-based methods, such as $LAMOL$ or $L2KD$ \cite{sun_lamol_2019, chuang_lifelong_2020}. In addition to the need for data sharing, prediction on new data may also be less efficient due to local adaptation strategies such as in $MbPA^{++}$ \cite{dautume_episodic_2019}, which can be a hindrance for the integration into running services. 

\subsection{Implementation Details}
For all conducted experiments, we build our code upon the \textit{Transformers} library \cite{wolf-etal-2020-transformers} and we use \textit{dmis-lab/biobert-base-cased-v1.1} as pre-trained model. For all experiments, due to lack of data, we did not perform hyperparameter optimization but train the models for three epochs, batch size of $16$ and with a learning rate of $3e-5$, except for the Adapter-based experiments, where we use a learning rate of $5e-4$. We provide our code under \url{https://github.com/llangnickel/WEAVER}. 

\subsection{Evaluation}
To evaluate our methods, we determine precision, recall and F1-score using the following formula (FP stands for false positive, FN for false negative and TP for true positive). 
\begin{equation}
precision=\frac{TP}{TP+FP} \quad recall=\frac{TP}{TP+FN} \end{equation}
\begin{equation}
F_1-score=2\times\left(\frac{precision\times recall}{precision+recall}\right)
\end{equation}

We determine the averaged precision, recall and F1-score over all test sets after the complete training procedure has been finished for all different training data set orders. In order to compare models, we use the Almost Stochastic Order test \cite{del2018optimal, dror2019deep} as 
implemented by \cite{ulmer2022deep}. Because the averaged F1-score determined at the end of training is not a sufficient measure to judge on the training performance, we, additionally, examine the extent of forgetting when re-training a model in a continual manner, by determining the Backward Transfer (BWT) according to \cite{lopez-paz_gradient_2022}. This metric measures the influence that learning a new task has on the previously learned task. Accordingly, we determine the Forward Transfer (FWT) that measures the influence of a learned task on the future tasks \cite{lopez-paz_gradient_2022}. The corresponding formula are depicted in Equations~\ref{eq:bwt} and \ref{eq:fwt}, respectively. Note, that, in contrast to \cite{lopez-paz_gradient_2022}, we use the F1-score instead of the accuracy score. Hence, $R \in R^{TxT}$ consists of the F1-scores on task $t_j$ after finishing training on task~$t_i$. 

\begin{equation}
    \label{eq:bwt}
    BWT = \frac{1}{T-1}\sum_{i=1}^{T-1} R_{T,i} - R_{i,i}
\end{equation}
\begin{equation}
    \label{eq:fwt}
    FWT = \frac{1}{T-1}\sum_{i=2}^{T} R_{i-1,i} - \overline{b}_{i}
\end{equation}

Moreover, we plot the extent of \emph{forgetting} by evaluating the performance on the test set corresponding to the training data set that has been used for the very first model after each re-training. 

\subsection{Visualization Techniques} 
To visualize the word (i.e. token) embeddings, we apply the dimensionality reduction technique Uniform Manifold Approximation and Projection (UMAP). More specifically, we make use of the python library umap-learn \cite{mcinnes2018umap-software}. 
This allows us to judge whether different data sets are embedded in different regions of the network or whether they share the representation space. Thereby, we compare the visualization obtained by the following settings: First, we train different models on different data sets individually, for example on the NCBI training and the BC5CDR training set in case of diseases NER. Then, we make predictions on these training data sets using the corresponding models and use the word embeddings which are vectors of length 768. Because this high dimensionality cannot be visualized, we apply UMAP to scale it down to two dimensions. We then color the embeddings of the different data sets (predicted by the two different models) differently. In addition, we use the baseline model that has been trained on both data sets simultaneously to also make predictions on both of these data sets. Finally, we visualize the word embeddings predicted by a model that has been trained sequentially on the mentioned data sets according to our developed method.
Since UMAP preserves cluster structures, this enables us to judge differences or overlaps of the embedding of different sets.

\subsection{Ablation Study}
Several studies show that end-to-end fine tuning is not necessary because only the final top layers are decisive for the downstream task. Accordingly, we freeze the first eight layers as suggested by \cite{lee_what_2019} and \cite{merchant_what_2020} to investigate whether weight averaging can be reduced to the top four layers.

\section{Results}\label{sec:results}
In the following section, we first describe the results of the models trained on a single corpus. Afterwards, the results for the simulated continual learning setting are given. Finally, we depict and discuss the UMAP visualizations of the word embeddings as proof-of-concept. The results of the performed ablation study can be found in the Appendix in Table~\ref{tab:ablation_study}. 

\subsection{Single Model Training and Cross-evaluation (Baseline)}
\begin{figure*}
    \centering
     \begin{subfigure}[b]{0.33\textwidth}
         \centering
         \includegraphics[width=1\textwidth]{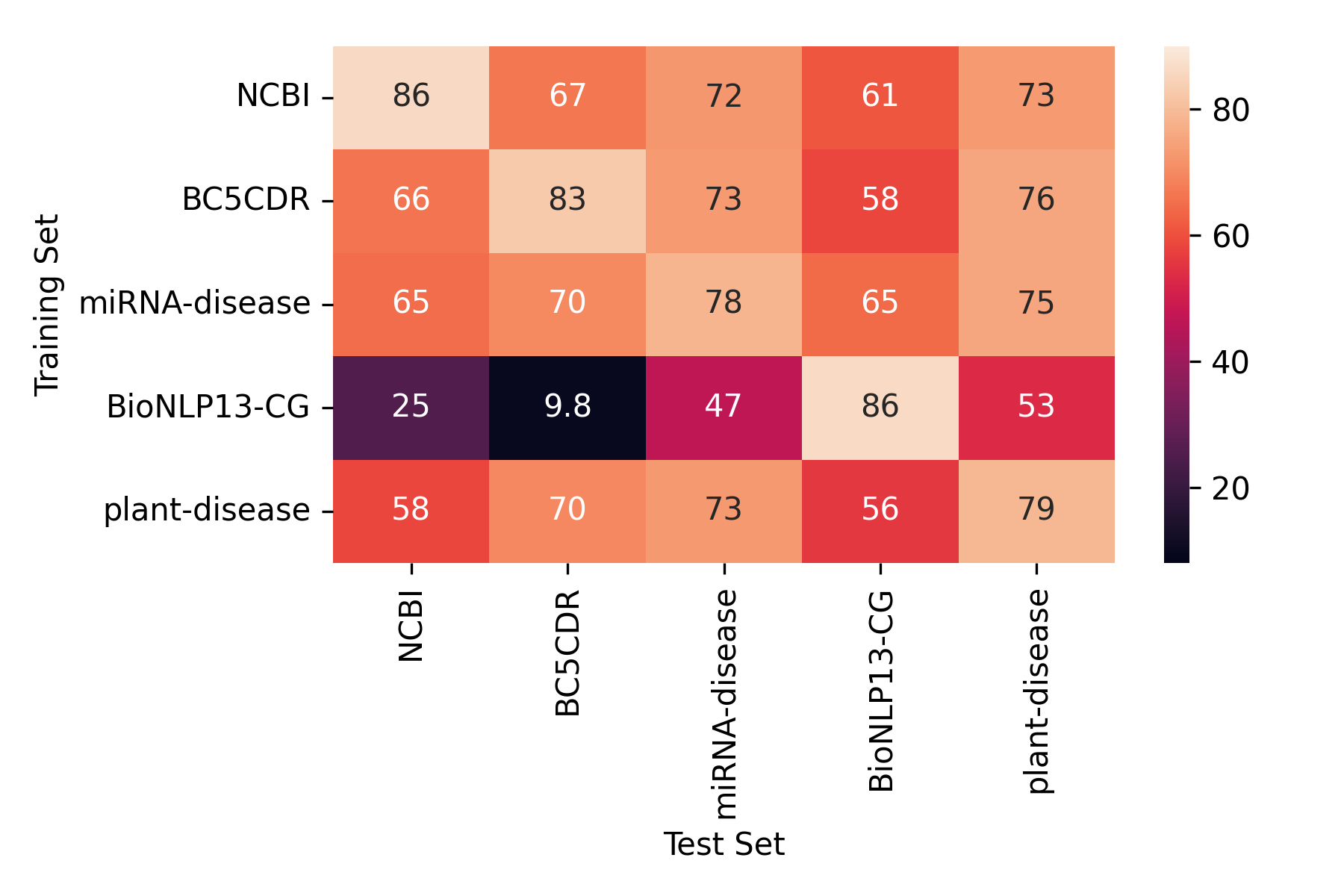}
         \caption{Diseases}
         \label{fig:single_models_diseases}
     \end{subfigure}
     \begin{subfigure}[b]{0.33\textwidth}
         \centering
         \includegraphics[width=1\textwidth]{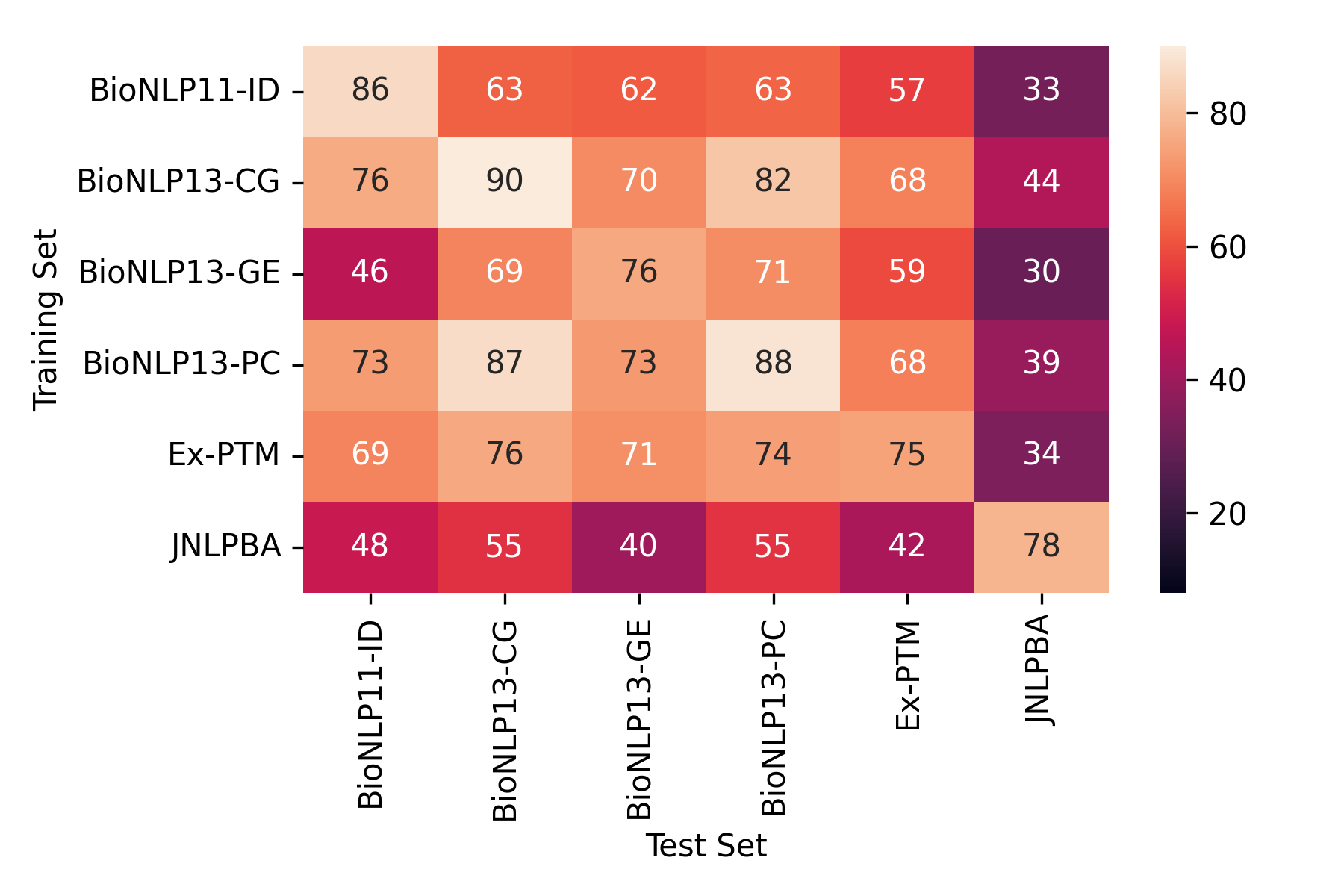}
         \caption{Genes/Proteins}
         \label{fig:single_models_proteins}
     \end{subfigure}
     \begin{subfigure}[b]{0.33\textwidth}
         \centering
         \includegraphics[width=1\textwidth]{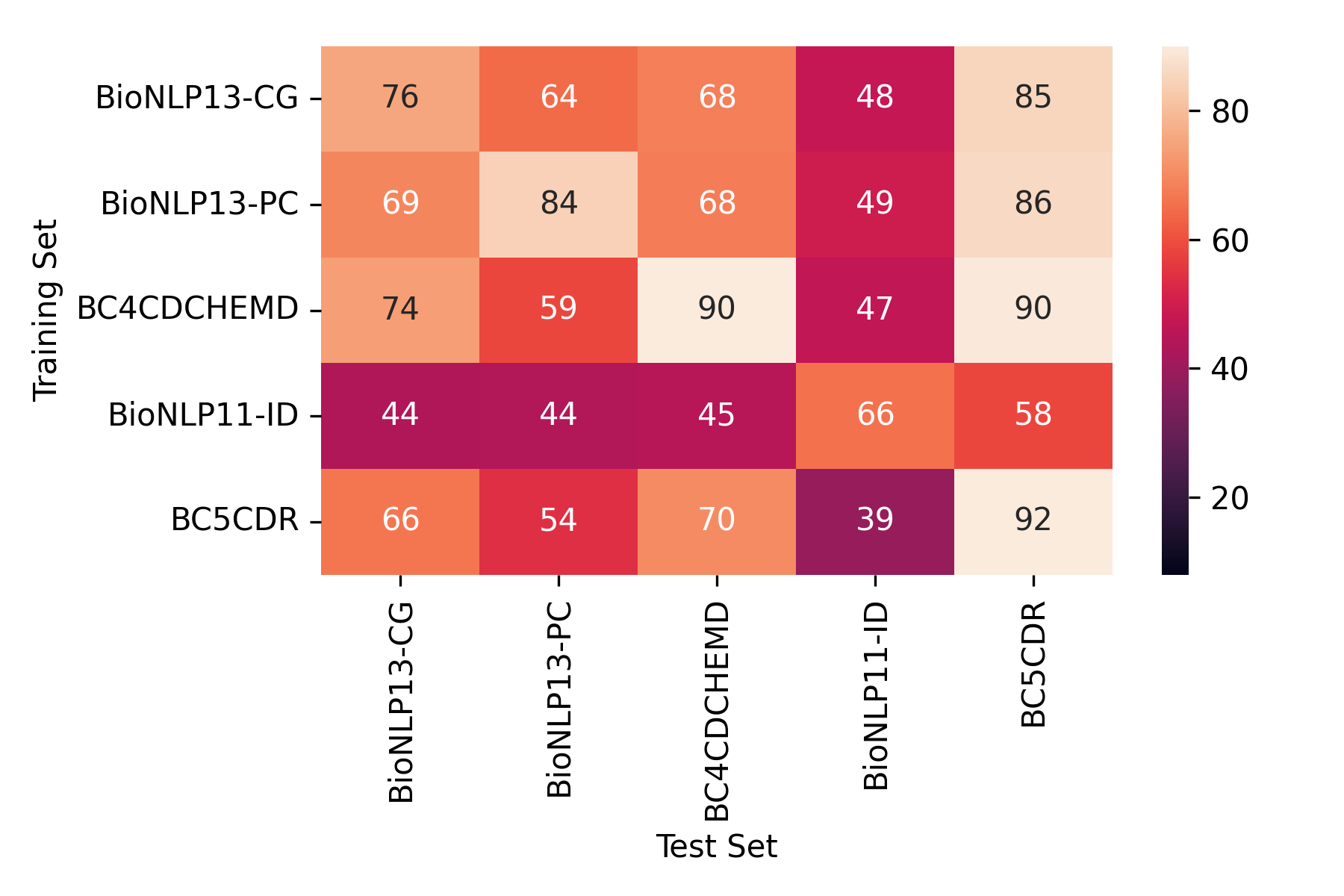}
         \caption{Chemicals}
         \label{fig:single_models_chemicals}
     \end{subfigure}
     \caption{Evaluation results of single models based on F1-scores. Each model has been trained on one of the available datasets and evaluated on all other datasets for the specific entity class.}
     \label{fig:single_model_evals}
\end{figure*}

\noindent For every entity class, several manually annotated data sets are available. As a first step, we train a BioBERT model on each of the data sets individually and evaluate the model on all available test sets for the given entity class. We visualize the result in a heatmap in Figure~\ref{fig:single_model_evals}. For each trained model, we see the highest score on the test set that belongs to the used training set and see significant drops in performance when evaluating the model on the other test sets. For example in case of disease entity recognition (see Fig.~\ref{fig:single_models_diseases}), the model trained on the NCBI disease corpus achieves an F1-score of 83\% on the corresponding test set, but drops to 66\% and 65\% for the BC5CDR and miRNA-disease corpus, respectively. The same phenomenon can be seen for all data sets across all three tasks (diseases, genes/proteins and chemicals). These results underline the fact that, even for the same task, trained models need to be improved as soon as new data sets exits because one available corpus may be too small or too specific. Hence, particularly for running services, continuous learning methods are of great importance.

\subsection{Continual Learning Experiments}
We simulate a continual learning scenario for three different named entity recognition use cases from the biomedical domain, namely diseases (5 data sets), genes/proteins (6 data sets) and chemicals (5 data sets), that are presented sequentially to the models without storing already seen data. We compare our developed method WEAVER with other continual learning-based methods (for transformers), namely FineTune, EWC and AdapterFusion. Additionally, we apply sparse memory replay (Replay) and multi-task learning (MTL) as upper bound methods because they require the data to be at the same place, which is not always possible in medical applications.

\par We apply four evaluation methods. First, the averaged F1-score on all available test data is determined after finishing training using the last training data set. The results are summarized in Figure~\ref{tab:averaged_f1}. For each entity class, four different orders of training data sets have been chosen randomly (see \ref{tab:orders}). For diseases, WEAVER outperforms all other CL methods. Whereas it achieves an average F1-score of 77.33\% for the different orders, with EWC and FineTune, we achieve 76.15\% and 76.59\%, respectively. For AdapterFusion, the difference is much higher; it achieves an F1-score of 63.51\%. In comparison to the upper bound MTL (79.50\%), WEAVER performs only around 2\% worse. For protein named entity recognition, WEAVER shows excellent results, that are on average only less than 1\% worse than the MTL model and outperforms the other CL-based methods. Interestingly, the Replay model does not work well here and achieves an average F1-score of 68\%, which could be caused by the fact that the replayed data is only learned for one epoch, which is probably not sufficient for these data sets. For chemical named entity recognition, a similar trend can be seen. WEAVER outperforms all other methods and is only around 1\% worse than MTL. For all three experiments, WEAVER shows on average the lowest standard deviation, thus the model's performance is less influenced by the training parameters/initialization. 

\begin{table*}
    \centering
    \caption{Averaged F1-scores after finishing training on all data sets. We averaged the F1-score over ten independent runs and determined the standard deviation (shown in brackets). Highest score is shown in bold, statistical significance has been tested using Almost Stochastic Order (see Table~\ref{tab:aso_values}). Upper-bound methods are shown on the right.}
    \begin{tabular}{cccccc|cc}
         Entity Class & Order* &  FineTune & EWC & AdapterFusion & WEAVER & Replay & MTL \\
         \hline 
         \multirow{4}{*}{Diseases}& (i) &  76.20 (0.44) & 74.24 (0.38) & 60.69 (2.34) & \textbf{76.77} (0.29) &78.09 (0.44) & 79.68 (0.29) \\
         & (ii) & 76.84 (0.23) & 77.13 (0.47) & 62.32 (2.87) & \textbf{77.36} (0.28) & 77.44 (0.89) & 79.49 (0.28) \\
         & (iii) & 76.63 (0.33) &  76.93 (0.49) & 62.09 (2.73) & \textbf{77.70} (0.18) & 77.33 (0.47) & 79.34 (0.25) \\
         & (iv) & 76.68 (0.48) & 76.29 (0.29) & 53.42 (11.93) & \textbf{77.47} (0.29) & 77.52 (0.64) & 79.48 (0.33) \\
         \hline
         & Avg. & 76.59 & 76.15 & 63.51 & \textbf{77.33} & 77.60 & 79.50 \\
         \hline
         \hline
         \multirow{4}{*}{Proteins}& (i) & 72.01 (0.32) & 72.28 (0.46) & 61.95 (1.3) & \textbf{75.60} (0.11) & 68.66 (1.09)
 & 76.05 (0.22) \\
         & (ii) & 75.84 (0.17) & \textbf{76.26} (0.41) & 63.66 (2.99) & 75.53 (0.17) & 68.44 (0.58) & 76.03 (0.22) \\
         & (iii) & 72.85 (0.38) & 71.91 (0.44) & 60.4 (2.39) & \textbf{74.43} (0.18) & 67.40 (0.55) & 75.98 (0.12) \\
         & (iv) & 73.34 (0.35) & 72.39 (0.39) & 60.29 (2.46) & \textbf{75.47} (0.16) & 67.14 (0.86) & 75.85 (0.28) \\
         \hline
          & Avg. & 73.51 & 73.21 & 61.58 & \textbf{75.26} & 67.91 & 75.98 \\
         \hline
         \hline
         \multirow{4}{*}{Chemicals}& (i) & 74.33 (0.72) & 73.70 (0.39) & 62.57 (2.05) & \textbf{76.81} (0.43) & 76.51 (0.51) & 78.26 (0.71) \\
         & (ii) & 74.27 (0.47) & 74.56 (0.51) & 62.08 (3.15) & \textbf{76.63} (0.13) & 76.22 (0.22) & 78.18 (0.12) \\
         & (iii) & 77.36 (0.42) & 77.54 (0.21) & 63.78 (4.69) & \textbf{77.76} (0.16) & 76.39 (0.69) & 78.27 (0.27) \\
         & (iv) & 75.05 (0.33) & 75.39 (0.32) & 64.32 (4.52) & \textbf{75.57} (0.36) & 75.90 (0.78) & 78.14 (0.43) \\
         \hline
         & Avg. & 75.25 & 75.30 & 63.19 & \textbf{76.69} & 76.26 & 78.21 \\
         \hline
         \hline
         \multicolumn{8}{l}{*See Table~\ref{tab:orders} for data set orders}
    \end{tabular}
    \label{tab:averaged_f1}
\end{table*}

As the averaged F1-score after finishing training does not indicate how training a new task influences both previous and future tasks, we additionally determine forward and backward transfer that are summarized in Table~\ref{tab:bwt_fwt}. Note, that as upper bound method only Replay can be used because for the multi-task setting, all training data are combined and no sequential training can be performed. Comparing the CL-based methods, WEAVER performs best for disease NER. Except for the first order of training data sets, we have a positive backward transfer, meaning that when learning a new task, the performance of a preceding task is increased. In two of the four cases, the BWT of WEAVER is also better than for Replay. In case of FWT, WEAVER achieves the best scores (approximately $0.5$). The forward transfer is positive for all scenarios, which is expected because we train the models on the same task sequentially, hence, learning on a specific data set will always positively influence a future task (as compared to random initialization of the model). In case of protein/gene NER, WEAVER achieves the best BWT scores for the CL-based methods, even though they are all slightly negative, meaning that learning on a new task results in moderate forgetting of the previously learned data. For the FWT, we achieve also scores around $0.5$ that are better than for Replay, but slightly worse than for FineTune. For example, for the first order, the FWT scores for FineTune and WEAVER amount to 0.4699 and 0.4655, respectively, indicating only a very small difference. For chemicals, we see a similar phenomenon. WEAVER achieves on average the highest FWT score, however, it is very similar for all methods, amounting to approximately $0.5$. Larger differences can be seen for the BWT, where WEAVER achieves for example a score of $-0.04$ for the first order and the value for FineTune is more than twice as bad ($-0.11$). 

\begin{table*}
    \centering
    \caption{Backward/Forward Transfer for all entity classes. Upper bound method Replay is shown on the right. Highest score is shown in bold. If highest score outperforms the upper-bound, it is also shown in italics.}
    \begin{tabular}{cccccc|c}
         Entity Class & Order* &  FineTune & EWC & AdapterFusion & WEAVER & Replay \\
         \hline 
         \multirow{4}{*}{Diseases}& (i) & -0.0918/0.5102 & -0.1103/0.509 & -0.1564/0.4845 & \textbf{-0.0020}/\textbf{\textit{0.5894}} & 0.0250/0.5441  \\
         & (ii) & -0.0775/0.4751 & -0.0754/0.4699 & -0.0674/0.4116 & \textit{\textbf{0.0732}}/\textbf{\textit{0.5061}} & 0.0549/0.4582 \\
         & (iii) & -0.0855/0.4949 & -0.0757/0.4979 & -0.0984/0.4384 & \textbf{\textit{0.0241}}/\textit{\textbf{0.5483}} & 0.0046/0.5157 \\
         & (iv) & -0.0909/0.5108 & -0.0968/0.5101 & -0.1865/0.4436 & \textbf{0.0312}/\textbf{\textit{0.5498}} & 0.0546/0.4245 \\
         \hline
         \multirow{4}{*}{Proteins}& (i) & -0.1411/\textbf{\textit{0.4699}} & -0.1376/0.4683 & -0.0891/0.4495 & \textbf{-0.0683}/0.4655 & 0.0786/0.4069 \\
         & (ii) & -0.1012/\textbf{\textit{0.5387}} & -0.0941/0.5393 & -0.1083/0.4834 & \textbf{-0.0708}/0.5304 & 0.0901/0.4214 \\
         & (iii) & -0.1386/\textbf{\textit{0.4863}} & -0.1463/0.4832 & -0.1616/0.4613 &\textbf{-0.0935}/0.4833 & 0.0611/0.4258 \\
         & (iv) & -0.1292/0.4931 & -0.135/0.4878 & -0.1407/0.4735 & \textbf{-0.0771}/\textit{\textbf{0.4963}} & 0.1025/0.4066  \\
         \hline
         \multirow{4}{*}{Chemicals}& (i) & -0.111/0.5147 & -0.1262/0.5113 & -0.1528/0.4827 & \textbf{-0.0436}/\textbf{\textit{0.5190}} & 0.0221/0.5025 \\
         & (ii) & -0.1174/0.5514 & -0.1161/\textbf{\textit{0.5453}} & -0.1357/0.5215 & \textbf{-0.0637}/0.5435 & 0.0235/0.4961  \\
         & (iii) & -0.0837/0.4669 & -0.0876/0.4617 & -0.0550/0.4267 & \textbf{-0.0374}/\textbf{0.4781} & 0.0439/0.4801 \\
         & (iv) & -0.1016/0.5282 & -0.0919/0.5236 & -0.0806/0.4889 & \textbf{-0.0792}/\textbf{\textit{0.5300}} &  0.0983/0.4333 \\
         \hline
         \multicolumn{7}{l}{*See Table~\ref{tab:orders} for data set orders}
    \end{tabular}
    \label{tab:bwt_fwt}
\end{table*}

\begin{figure*}
    \centering
     \begin{subfigure}[b]{0.33\textwidth}
         \centering
         \includegraphics[width=1\textwidth]{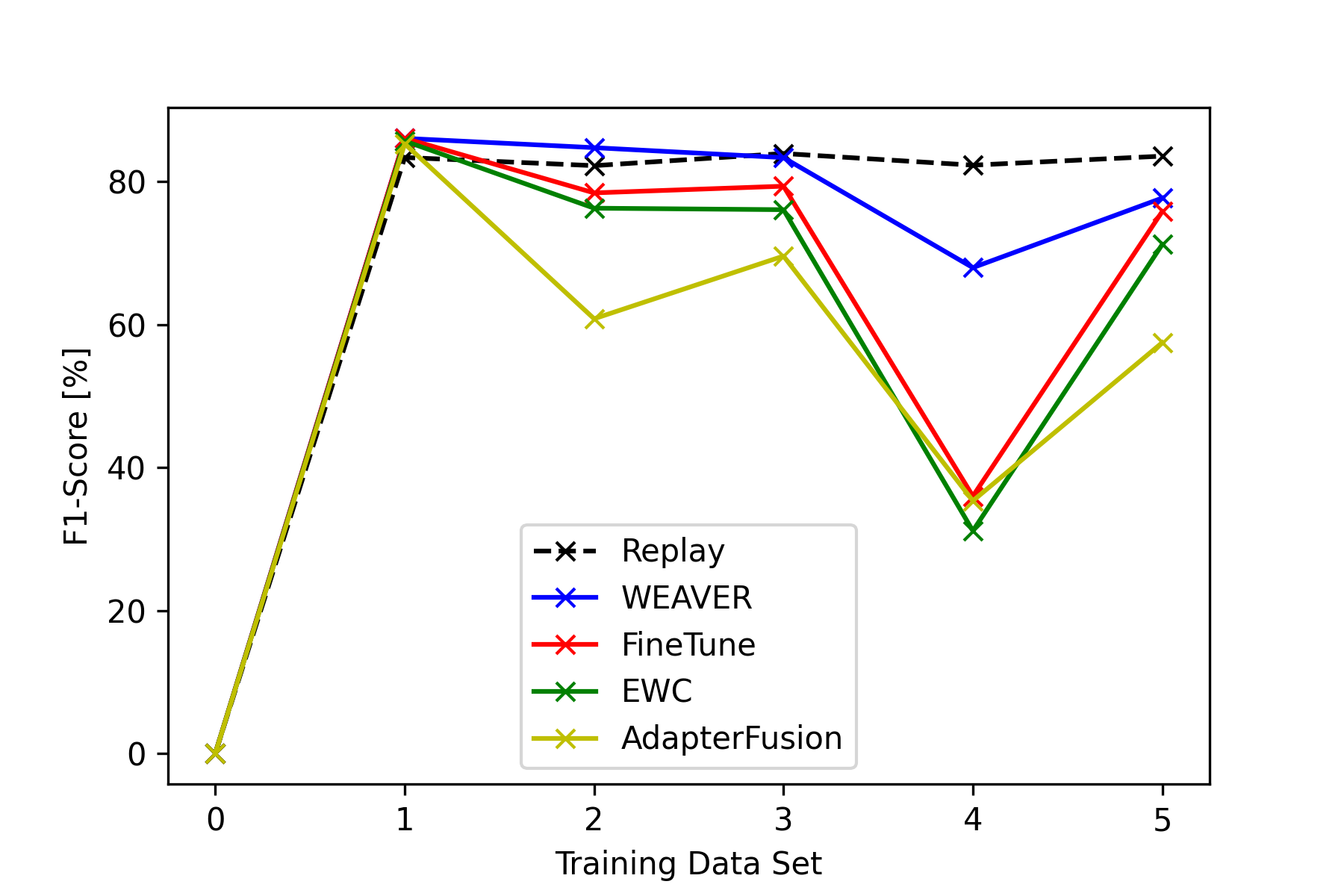}
         \caption{Diseases}
         \label{fig:forgetting_diseases}
     \end{subfigure}
     \begin{subfigure}[b]{0.33\textwidth}
         \centering
         \includegraphics[width=1\textwidth]{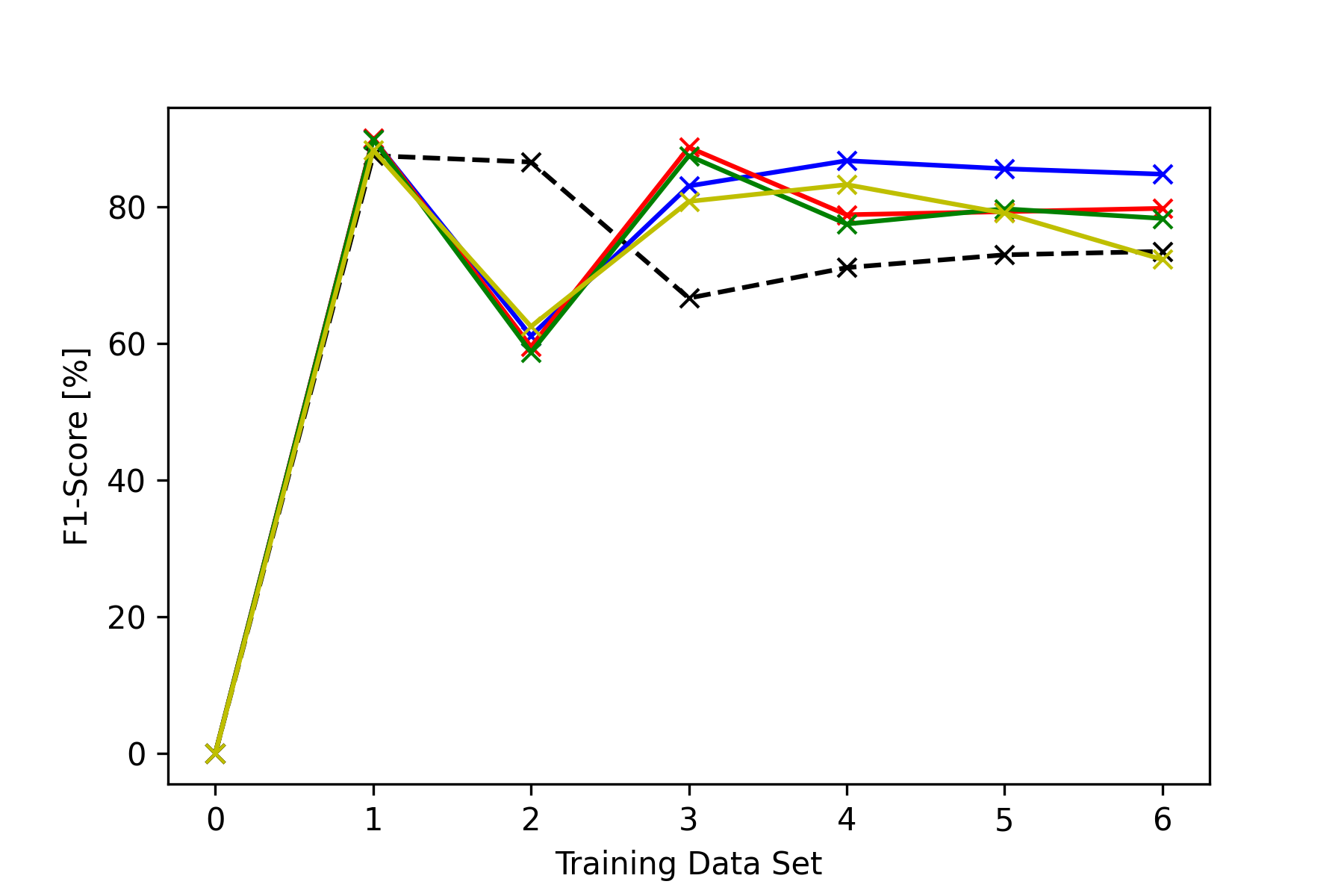}
         \caption{Genes/Proteins}
         \label{fig:forgetting_proteins}
     \end{subfigure}
     \begin{subfigure}[b]{0.33\textwidth}
         \centering
         \includegraphics[width=1\textwidth]{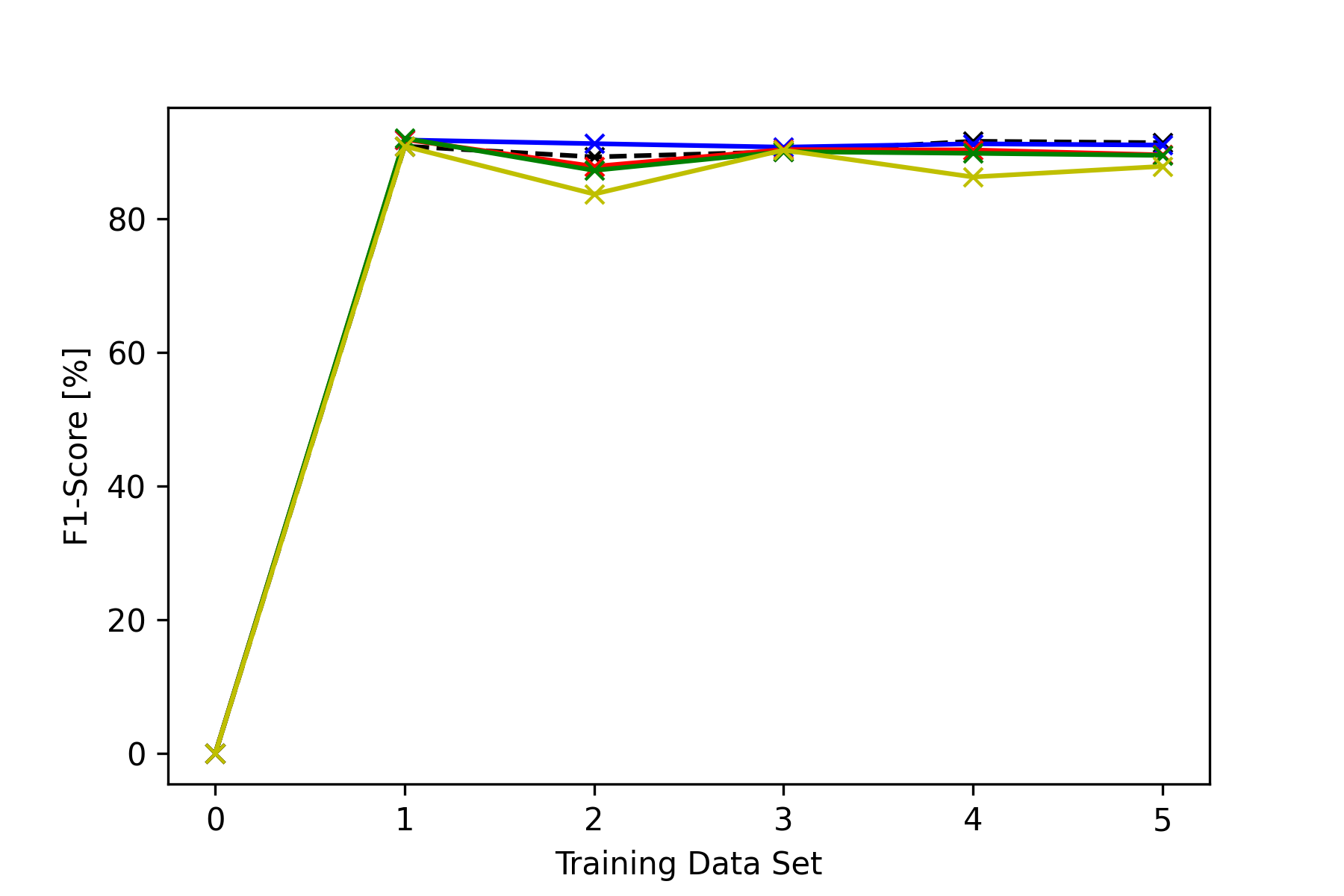}
         \caption{Chemicals}
         \label{fig:forgetting_chemicals}
     \end{subfigure}
     \caption{\textbf{F1-scores on the first test data set over time.} After each re-training of the model, it is evaluated on the test set corresponding to the first training data set in order to see how much the model forgets. The legend depicted in subfigure (a) equally applies to the two other subfigures.}
     \label{fig:forgetting}
\end{figure*}

As a last evaluation metric, we plot the extent of forgetting in Fig.~\ref{fig:forgetting}. Thereby, we determined the F1-score for the test set that corresponds to the very first training data set after random initialization and after each re-training of the model in order to see how much the model "forgets" when being exposed to new data. For the disease NER (see Fig.~\ref{fig:forgetting_diseases}), FineTune, EWC and AdapterFusion drop to an F1-score of around 30\%, WEAVER only drops to around 60\% after seeing the fourth training data set. Also after finishing the last training, WEAVER only performs slightly worse than Replay. For protein NER, the CL-based methods perform very similar until the fourth training data set, where the highest score can now be seen for WEAVER. It also outperforms Replay, which which goes in line with the average performance seen in Table~\ref{tab:averaged_f1}. For Chemicals, we also achieve the highest F1-score with WEAVER, however, the difference to all other methods is rather small in this case. 

\subsection{Visualization of Word Embeddings}
\begin{figure*}[h!]
     \centering
     \captionsetup[subfigure]{justification=centering}
     \begin{subfigure}[b]{0.3\textwidth}
         \centering
         \includegraphics[width=1\textwidth]{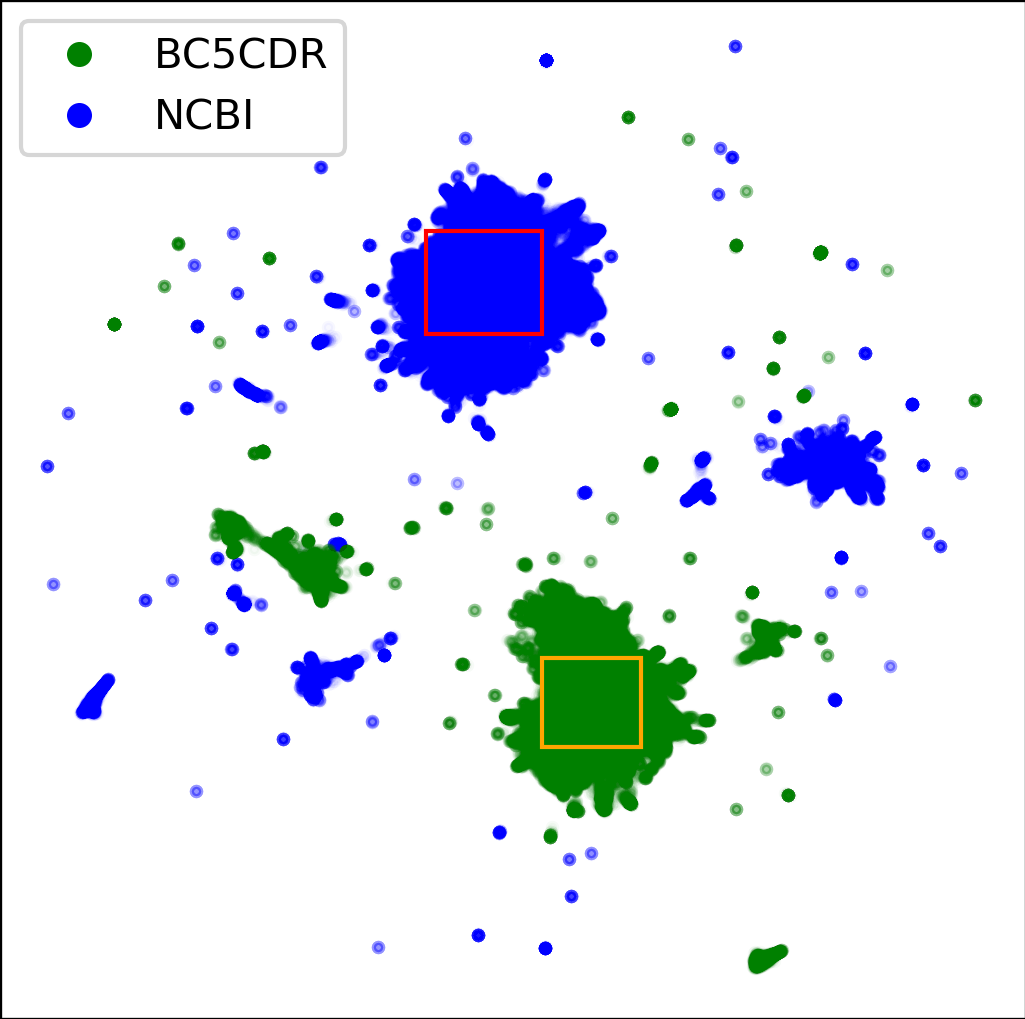}
         \caption{NCBI and BC5CDR models \newline trained independently}
         \label{fig:umap_trained_independently}
     \end{subfigure} %
     \begin{subfigure}[b]{0.3\textwidth}
         \centering
         \includegraphics[width=1\textwidth]{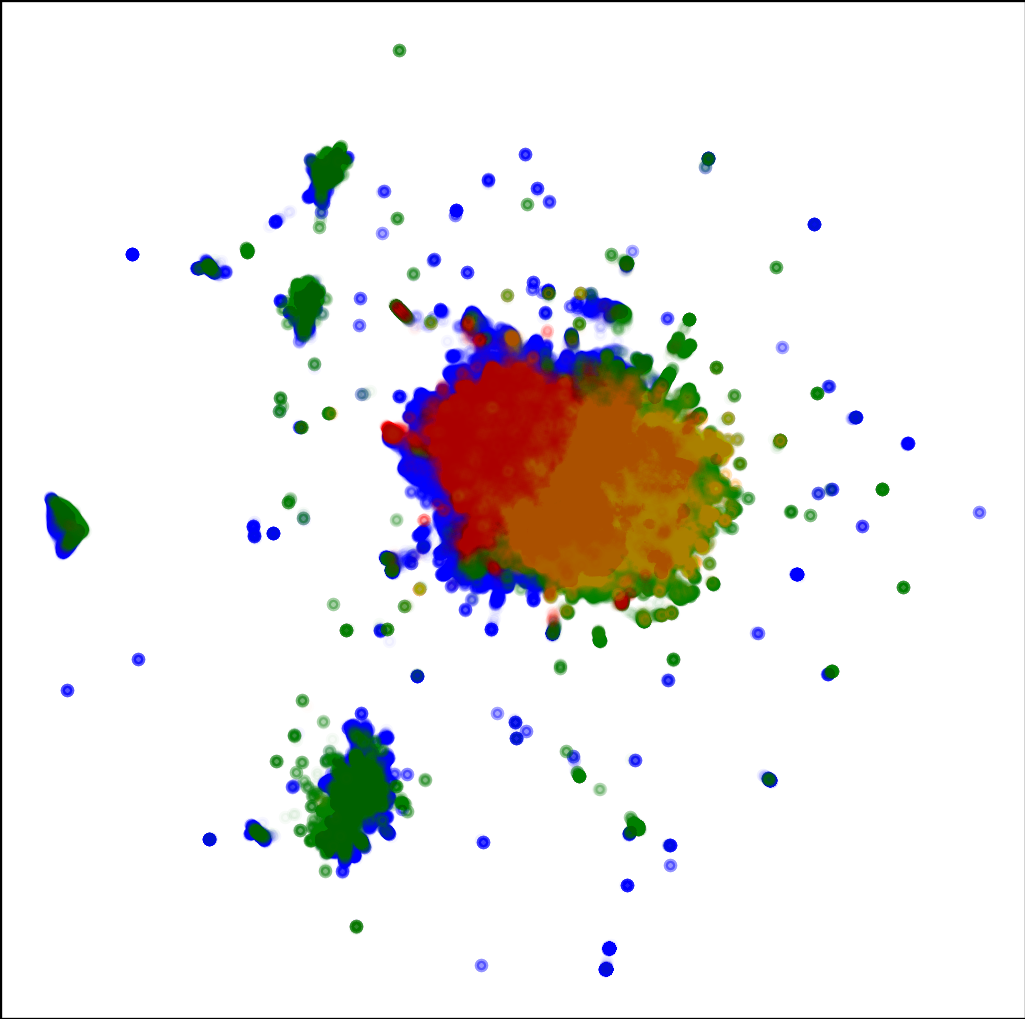}
         \caption{NCBI and BC5CDR models \newline trained jointly}
         \label{fig:umap_trained_jointly}
     \end{subfigure}
     \begin{subfigure}[b]{0.3\textwidth}
         \centering
         \includegraphics[width=1\textwidth]{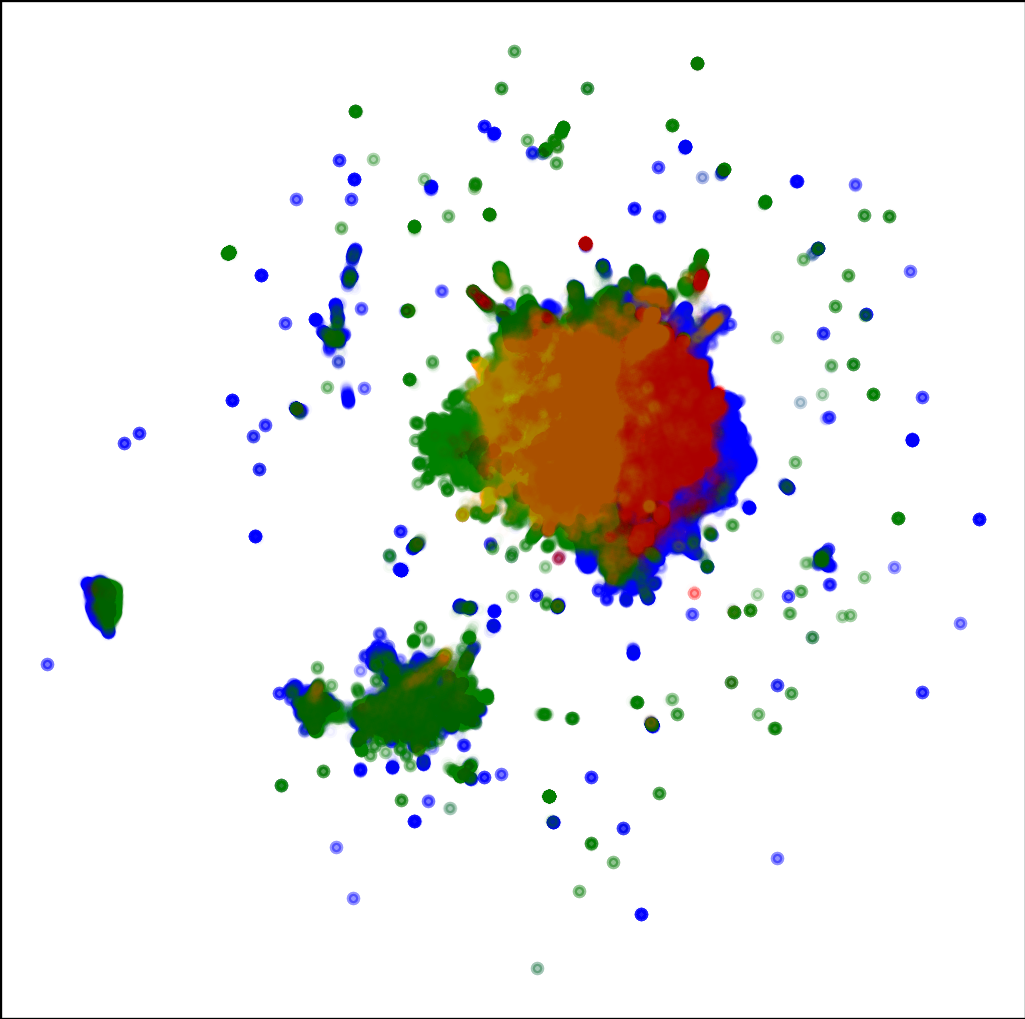}
         \caption{Model trained continually on NCBI and BC5CDR using WEAVER}
         \label{fig:umap_trained_with_weaver}
     \end{subfigure}
    \begin{subfigure}[b]{0.3\textwidth}
         \centering
         \includegraphics[width=1\textwidth]{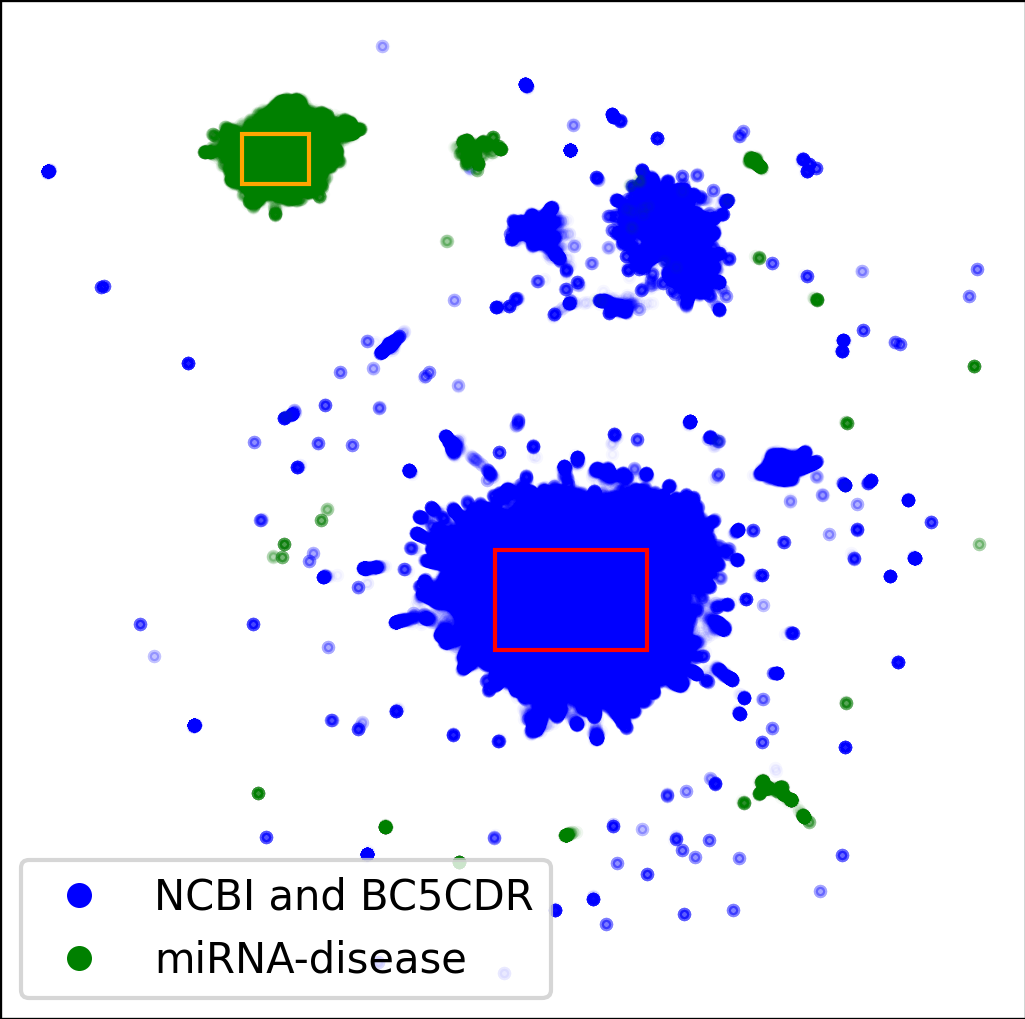}
         \caption{Model trained independently on NCBI+BC5CDR and miRNA-disease \newline data set}
         \label{fig:umap_trained_independent_2}
     \end{subfigure}
     \begin{subfigure}[b]{0.3\textwidth}
         \centering
         \includegraphics[width=1\textwidth]{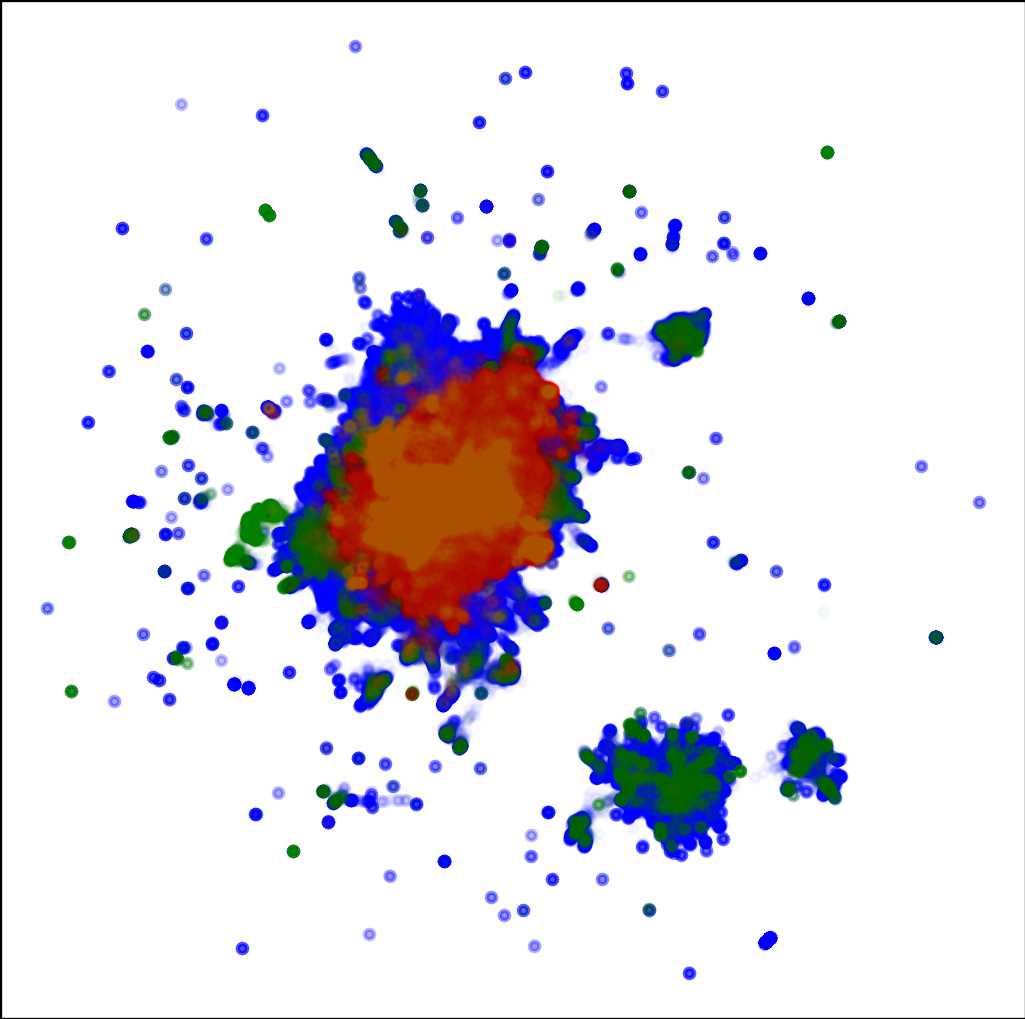}
         \caption{Model trained jointly on NCBI, BC5CDR and miRNA-disease \newline data set}
         \label{fig:umap_trained_jointly_2}
     \end{subfigure}
     \begin{subfigure}[b]{0.3\textwidth}
         \centering
         \includegraphics[width=1\textwidth]{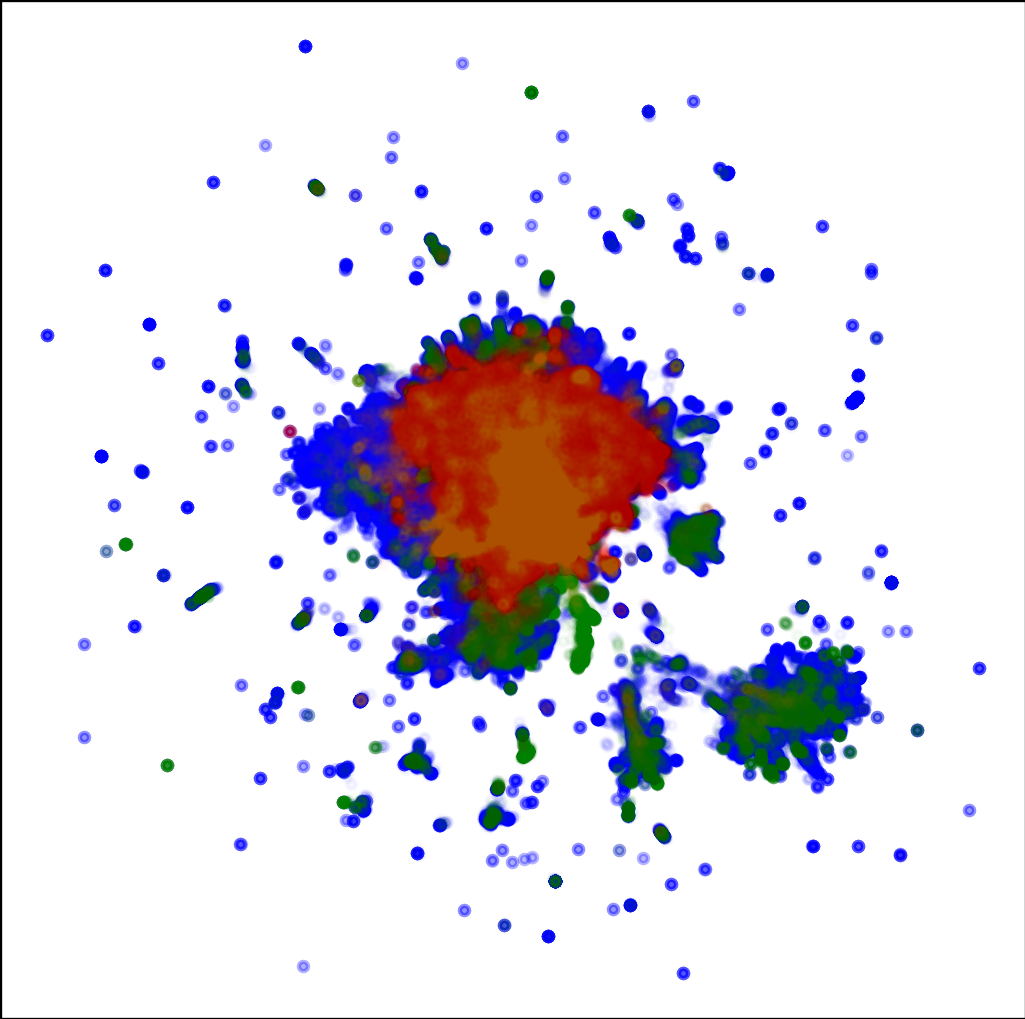}
         \caption{Model trained continually on NCBI, BC5CDR and miRNA-disease using WEAVER}
         \label{fig:umap_trained_with_weaver_2}
     \end{subfigure}
        \caption{\textbf{BERT embeddings visualized using UMAP.} The different sub-figures show the distribution of the word embeddings predicted for three different disease NER data sets (NCBI, BC5CDR and miRNA-disease) using different models. In sub-figure (a), two different models were used that have been trained independently on the two data sets. In contrast, predicted word embeddings from a model trained on the combined training data are depicted in (b). In sub-figure (c), the embeddings resulting from the continually trained model using WEAVER is shown. With the red and yellow squares depicted in (a), we show where the corresponding word embeddings moved to in settings (b) and (c). In subfigures d-e, the same setting is shown but now for a third data set. Therefore, the previously used data sets NCBI and BC5CDR are represented in one color.}
        \label{fig:umap_diseases}
\end{figure*}

In order to comprehend what happens to the word embeddings when averaging the weights of two BERT models, we performed a UMAP visualization for the different scenarios that can be seen in Figure~\ref{fig:umap_diseases}. Exemplary, we use the disease NER use case and compare the arrangement of the embeddings for three different training sets (NCBI, BC5CDR and miRNA-disease) to simulate continual training. Therefore, first, two models have been trained independently on the NCBI and BC5CDR data set, respectively, and their predicted embeddings are visualized in Fig.~\ref{fig:umap_trained_independently}. Embeddings for the different data sets, predicted by the two different models are clearly separated. In contrast, in Fig.~\ref{fig:umap_trained_jointly}, where a model trained on both data sets simultaneously is used for prediction, the points are strongly overlapping and separate clusters cannot be recognized. Figure~\ref{fig:umap_trained_with_weaver} shows word embeddings predicted by a model trained according to our method WEAVER (first NCBI training set, then BC5CDR training set). Interestingly, it can be seen that the distribution looks very similar to a combined training. Thereby, we can infer that weight averaging after training two models sequentially has a similar effect to a combined training (simultaneously on all training data). To investigate that effect when training on more than two data sets, we use a third one (miRNA-disease data set). Figures~\ref{fig:umap_trained_jointly_2}-\ref{fig:umap_trained_with_weaver_2} visualize the same settings as described before but now, the first two data sets (NCBI and BC5CDR) are combined to one color so that the new data set can be clearly distinguished. Here, we see the same phenomenon, i.e. that training sequentially with WEAVER results in very similar distributions of the embeddings than training one model jointly on all three data sets.

\section{Discussion}\label{sec:discussion}
Transformer-based models have boosted the advancements of natural language processing tasks. Especially in the biomedical domain, BERT-based models are adapted to specific applications, such as electronic health record (EHR) mining or the identification of rare disease patients from administrative claims \cite{rasmy_med-bert_2021, li_behrt_2020, prakash_rarebert_2021}. However, these models have the underlying assumption that data are independent and identically distributed. In real world scenarios in the biomedical domain, this is unfortunately not the case, in particular since current models do not yet represent all facets, which are present in such corpora due to much smaller training sets as compared to general domains. Hence different corpora easily display novel and different aspects here, which correspond to a shift of the distribution. 

In previous studies, we showed that there are significant differences in different data sets and that a model trained on one corpus does not perform well on another corpus; i.e., one such annotated corpus is not representative for biomedical literature data bases such as PubMed. Therefore, to be used in real world applications, trained models need the ability of lifelong learning (also on the same task) - meaning that they can be improved continuously without suffering from catastrophic forgetting. Whereas a lot of research has been done in this direction, most of the approaches do either need also previous data when training on the new data (i.e. (pseudo-)rehearsal), consists of a more complex structure containing two or more different networks (i.e. for example a knowledge base and an active column) or are, in case of regularization-based methods, computationally more inefficient.

\par Therefore, we propose a lifelong learning algorithm that (1) is based on transformers as current state-of-the-art methods, (2) can be used for federated learning if the data sets are not available at one place (e.g. in clinical use cases due to data privacy), (3) does not involve a second or different neural network structure hence requires limited resources, and (4) is computationally efficient.    

\par We evaluated our method on three different use cases from the biomedical domain, namely diseases, genes/proteins and chemicals. For these entity classes five or six different data sets are available, respectively (see Table~\ref{tab:overview_datasets}). As baseline, we first determined the evaluation results on single models, i.e. models that have been trained on a single training data set. These are evaluated on the corresponding test set but also on all other test sets from this domain (called cross-evaluation). Here, we see significant differences (compare Figure~\ref{fig:single_model_evals}). For example, a model trained on the NCBI disease corpus performs best on its available test set (F1-score of 86\%) but drops to 25\% for the BioNLP13-CG data set that focuses on cancer related diseases. Similar is true for genes and chemicals where the F1-score can differ about 50\%. Hence, continual improvement of the models is needed.

\par We compared our method WEAVER to several different transformer-based CL methods that, except for Replay, do not require the data to be at the same place. We show that WEAVER outperforms the other methods in terms of average F1-score after finishing training of the last data set (see Table~\ref{tab:averaged_f1}), backward and forward transfer (Table~\ref{tab:bwt_fwt}) as well as on the performance of the test set corresponding to the very first data set (Fig.~\ref{fig:forgetting}) for all three use cases. However, the evaluation turns out differently for the different use cases. In terms of averaged F1-score, for disease NER, WEAVER is for example about 1\% better than FineTune and EWC, and around 2\% worse than the upper bound (MTL). In case of protein NER, WEAVER is only less than 1\% worse than MTL and around 2\% better than FineTune and EWC. In all scenarios, AdapterFusion performs worst. 

\par In terms of disease NER, WEAVER achieves mainly positive backward transfer and outperforms all other CL-based methods. Generally, for backward transfer, we see differences between the different orderings, indicating that the order can influence the success of training. For the forward transfer, this is less noticeable, for all use cases and orderings, the values range from $0.4$ to $0.5$. 

\par As we plotted the extent of forgetting in Fig.~\ref{fig:forgetting}, it can be seen that WEAVER is more robust to training of new data sets, i.e. it shows less variation in its performance of the test set corresponding to the very first training data set when seeing a new data set. 

\par For proof of concept of WEAVER, we visualized token embeddings of the variously trained models. Figure~\ref{fig:umap_diseases} indicates that applying WEAVER to a series of new data sets results in similar word embedding distributions as the combined training - with the advantage of efficiently improving a model as soon as new data sets arise. 

Summarizing, WEAVER consists of only one small post processing step where weights are averaged. In comparison to other presented methods, there is no need to change the training procedure; in addition, this method can theoretically not only be applied to transformer-based methods but to all neural network-based methods where weights are determined by training. However, possible limitations of our proposed method need to be further investigated: Since the averaging is weighted based on the size of the training data sets, this can be dangerous if sizes differ too much. For example, if a model is re-trained on a big data set which only represents a small sub-domain (e.g. cancer related diseases), the model can be still biased towards this data set/topic. Therefore, further experiments are needed to investigate the influence and importance of weighting based on the corpus size. Thereby, the previous recognition of a shift could also be useful and needs to be incorporated into future experiments \cite{DBLP:conf/ideal/FeldhansWHSHNH21}. Still, WEAVER shows very good results and outperforms other CL-based methods that do not require the data to be at the same place. Thereby, it perfectly combines practicability and quality, and can hence also be used for the continuous improvement of running services, such as our semantic search engine preVIEW COVID-19 \cite{langnickel_preview_2021, langnickel_covid-19_2021}. 

\section{Conclusion}\label{sec:conclusion}
Based on transformer models as state-of-the-art methods for document processing, we propose a new lifelong learning method called WEAVER. This method continuously trains a model on top of a previously trained model and infuses knowledge from the previously trained model into the new one by weight averaging. Thereby, we demonstrate a simple, yet efficient method that can also be used in settings, where the data sets are not available in one place. This is especially important in clinical use cases where data sets underlie data protection laws. In addition, in contrast to conventional federated learning settings, no central server is needed but the weights can be simply passed from one institution to the next. Moreover, our method is a simple post-processing step, which means that the training workflow itself does not need to be changed and therefore it can be easily implemented into running services, such as semantic search engines. Therefore, in future work, the method will be tested on other NLP tasks from the biomedical domain, such as document classification, and will be integrated into our semantic search engine \cite{langnickel_preview_2021}. 

\bibliographystyle{unsrt}  
\bibliography{references}  

\newpage
\section*{Appendix}

\setcounter{table}{0}
\renewcommand{\thetable}{A\arabic{table}}
Orderings:
\begin{table}[h!]
    \centering
    \caption{Overview of randomly chosen orderings of the data sets.}
    \begin{tabular}{lll}
         \textbf{Entity Class} & \multicolumn{2}{l}{\textbf{Order}} \\
         \hline
         \multirow{4}{*}{Disease} & (i) & NCBI $\rightarrow$ miRNA-disease $\rightarrow$ BC5CDR $\rightarrow$ BioNLP13-CG $\rightarrow$ plant-disease \\
         & (ii) & plant-disease $\rightarrow$ BioNLP13-CG $\rightarrow$ miRNA-disease $\rightarrow$  BC5CDR $\rightarrow$ NCBI \\
         & (iii) & BC5CDR $\rightarrow$ BioNLP13-CG $\rightarrow$ miRNA-disease $\rightarrow$ plant-disease $\rightarrow$ NCBI \\
         & (iv) & BioNLP13-CG $\rightarrow$ plant-disease $\rightarrow$ NCBI $\rightarrow$ miRNA-disease $\rightarrow$ BC5CDR \\
         \hline
         \multirow{4}{*}{Proteins} & (i) & BioNLP11-ID $\rightarrow$ BioNLP13-GE $\rightarrow$ JNLPBA $\rightarrow$ BioNLP13-CG $\rightarrow$ BioNLP13-PC $\rightarrow$ Ex-PTM \\
         & (ii) & JNLPBA $\rightarrow$ BioNLP11-ID $\rightarrow$ BioNLP13-CG $\rightarrow$ Ex-PTM $\rightarrow$ BioNLP13-GE $\rightarrow$ BioNLP13-PC \\
         & (iii) & BioNLP13-CG $\rightarrow$ JNLPBA $\rightarrow$ BioNLP13-PC $\rightarrow$ BioNLP13-GE $\rightarrow$ Ex-PTM $\rightarrow$ BioNLP11-ID \\
         & (iv) & Ex-PTM $\rightarrow$ JNLPBA $\rightarrow$ BioNLP13-PC $\rightarrow$ BioNLP13-GE $\rightarrow$ BioNLP13-CG $\rightarrow$ BioNLP11-ID \\
         \hline
         \multirow{4}{*}{Chemicals} & (i) & BioNLP11-CG $\rightarrow$ BC4CHEMD $\rightarrow$ BioNLP13-PC $\rightarrow$ BioNLP11-ID $\rightarrow$ BC5CDR-chem \\
         & (ii) & BioNL13-PC $\rightarrow$ BC4CHEMD $\rightarrow$ BC5CDR-chem $\rightarrow$ BioNLP13-CG $\rightarrow$ BioNLP11-ID \\
         & (iii) & BC5CDR-chem $\rightarrow$ BioNLP11-ID $\rightarrow$ BC4CHEMD $\rightarrow$ BioNLP13-PC $\rightarrow$ BioNLP13-CG \\
         & (iv) & BioNLP11-ID $\rightarrow$ BioNLP13-PC $\rightarrow$ BC4CHEMD $\rightarrow$ BC5CDR-chem $\rightarrow$ BioNLP13-CG \\
         \hline
    \end{tabular}
    \label{tab:orders}
\end{table}

We compared WEVAER to all other CL models based on ten random seeds each using Almost Stochastic Order with a confidence level of $\alpha = 0.05$. Almost stochastic dominance ($\epsilon_{min} < \tau$ with $\tau = 0.2$) is indicated in Table~\ref{tab:aso_values}.

\begin{table}[h!]
    \centering
    \caption{$\epsilon_{min}$ values to compare WEAVER with all other models using Almost Stochastic Order with $\tau = 0.2$}
    \begin{tabular}{cclll|l}\label{tab:aso_values}
         Entity Class & Order & FineTune & EWC & AdapterFusion & Replay  \\
         \hline
        \multirow{4}{*}{Diseases} & (i) & 0.0532 & 0.0 & 0.0 & 0.9971 \\
         & (ii) & 0.0005 & 0.0912 & 0.0 & 0.9514 \\
         & (iii) & 0.0 & 0.0 & 0.0 & 0.1146 \\
         & (iv) & 0.0065 & 0.0 & 0.0 & 1.0 \\
         \hline
         \multirow{4}{*}{Proteins} & (i) & 0.0 & 0.0 & 0.0 & 0.0 \\ 
         & (ii) & 0.9971 & 0.9932 & 0.0 & 0.0 \\
         & (iii) & 0.0 & 0.0 & 0.0 & 0.0 \\ 
         & (iv) & 0.0 & 0.0 & 0.0 & 0.0 \\
         \hline
         \multirow{4}{*}{Chemicals} & (i) & 0.0 & 0.0 & 0.0 & 0.6043 \\
         & (ii) & 0.0 & 0.0 & 0.0 & 0.6709 \\ 
         & (iii) & 0.2046 & 0.0966 & 0.0 & 0.0 \\ 
         & (iv) & 0.1175 & 0.7511 & 0.0 & 1.0 \\
         \hline
    \end{tabular}
    \label{tab:my_label}
\end{table}

For the ablation study, we freeze the first eight layers and only fine-tune the last four. As can be seen in Table \ref{tab:ablation_study}, this results in lower F1-scores than fine-tuning the whole model for WEAVER. 

\begin{table}[h!]
    \centering
    \caption{Comparison of averaged F1-scores (over ten runs) on test sets during training with different amounts of frozen layers. Standard deviation is shown in brackets.}
    \begin{tabular}{lclllll}
         & & \multicolumn{5}{c}{\textbf{Training data set}} \\
        \textbf{Fine-tuned layers} & \textbf{Order} & \textbf{1} & \textbf{2} & \textbf{3} & \textbf{4} & \textbf{5} \\
        \hline
        All & \multirow{2}{*}{(i)} & 71.94 (0.22) & 74.78 (0.39) & 77.54 (0.32) & 67.66 (0.59) & 76.77 (0.29)\\
        Last four & & 68.36 (0.41) & 70.57 (0.19) & 72.87 (0.15) & 63.78 (0.38) & 72.64 (0.18) \\
        \hline
        All & \multirow{2}{*}{(ii)} & 67.01 (0.78) & 47.91 (0.72) & 69.92 (0.66) & 75.76 (0.26) & 77.36 (0.28) \\
        Last four & & 61.09 (0.58) & 46.48 (0.38) & 63.31 (0.59) & 71.7 (0.14) & 73.64 (0.27) \\
        \hline
        All & \multirow{2}{*}{(iii)} & 71.15 (0.26) & 62.08 (1.23) & 73.67 (0.56) & 75.46 (0.37) & 77.70 (0.18)\\
        Last four & & 67.34 (0.48) & 58.47 (0.36) & 69.89 (0.33) & 72.24 (0.19) & 74.1 (0.16) \\
        \hline
        All & \multirow{2}{*}{(iv)} &44.15 (0.19) & 62.74 (0.88) & 74.96 (0.21) & 76.16 (0.38) & 77.47 (0.29) \\
        Last four & & 42.98 (0.06) & 57.03 (0.33) & 70.49 (0.21) & 72.16 (0.21) & 73.59 (0.15) \\
        \hline
        \hline
        All & \multirow{2}{*}{Avg.} & 63.56 & 61.88 & 74.02 & 73.76 & 77.32 \\
        Last four & & 59.94 & 58.14 & 69.14 & 69.97 & 73.50 \\
        \hline
    \end{tabular}
    \label{tab:ablation_study}
\end{table}

\end{document}